\documentclass[letterpaper, 10 pt, conference]{ieeeconf}
\IEEEoverridecommandlockouts
\usepackage[backend=biber, style=ieee]{biblatex}
\usepackage[ruled,vlined]{algorithm2e}
\usepackage{amsmath}
\usepackage{graphicx}
\usepackage{fullpage}
\usepackage{subcaption}
\usepackage{url}
\usepackage{csquotes}
\usepackage[american]{babel}
\usepackage{comment}
\usepackage{multirow}

\newtheorem{theorem}{Theorem}
\newtheorem{lemma}[theorem]{Lemma}

\newenvironment{myprocedure}[1][htb]
  {
   \begin{algorithm}[#1]%
  }{\end{algorithm}}

\title{Optimizing towards the best insertion-based error-tolerating joints}
\author{
Zhibin Zou \\
Department of Electrical and Computer Engineering\\ 
University at Albany SUNY, Albany, NY 12222 USA\\
zzou2@albany.edu 
}
\date{}

\begin{document}
\maketitle

\begin{abstract}
We present an optimization-based design process that can generate the {\em best} insertion-based joints with respect to different errors, including manipulation error, manufacturing error, and sensing error. We separate the analysis into two stages, the {\em insertion} and the {\em after-insertion stability}. Each sub-process is discretized into different {\em mode of contacts}. The transitions among the contact modes form a directed graph, and the connectivity of the graph is achieved and maintained through the manipulation of the socket edge-angle and peg contact-point locations. The analysis starts in 2D with the assumption of point-edge contacts. During the optimization, the edges of the socket are rotated and the points on the peg are moved along the edges to ensure the successful insertion and the stability after insertion. We show in simulation that our proposed method can generate insertion-based joints that are tolerant to the given errors. and we present a few simple 3D projections to show that the analysis is still effective beyond 2D cases. 
\end{abstract}

\section{Introduction}

In this work, we present algorithms to automatically design insertion-based (peg-in-hole) joints that allow easy insertion and maintains stability after insertion, subject to sensing, manipulation, and manufacturing errors. The peg-in-hole problem was studied as a theoretical problem a couple decades ago~\cite{lozano1984automatic}, and lead to back-chaining based approaches that can be applied to many different scenarios. Practically, however, even with state-of-the-art sensing technologies and robots, it is still difficult to insert tight joints, like Lego. Recently, there have been works on how to learn insertion strategies for tight joints~\cite{park2017compliance, yun2008compliant}. There also has been many studies on how to design and engineer a peg-in-hole type joint for given application and specifications~\cite{bruyninckx1995peg}. 

Insertion-based joints are widely used in many modern assembly tasks. At the same time, the problem is yet to be solved in general practically. Insertion task is a prerequisite for many modern assembly tasks. For example, screws need to be aligned and inserted before it can be successfully twisted into the socket. One of the main observations is that the compliance and the capability to adapt to the contact forces are important to the success of tight joint insertion. It is still difficult to introduce adaptability and compliance to the modern robot arms. The design process we propose in this work attempts to use good geometric designs to {\em simulate} adaptive strategies, and mitigate the responsibility of compliance to the joint and socket designs. 

\begin{figure}[h]
\begin{center}
\includegraphics[width=3in]{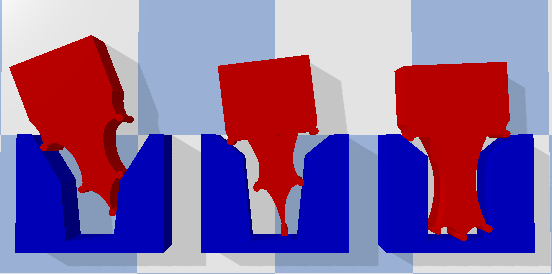}
\end{center}
\caption{During insertion, arbitrary design may get {\em stuck} before reaching the sink, while the optimized joint can reach the sink regardless of the initial condition. }
\label{fig:stuck}
\end{figure}

We present an automated design procedure to find {\em best} feasible design for insertion-based joints that can overcome different practical challenges. By best, we mean a design that is most stable after insertion while can still be successfully inserted into the socket. In a generic setting, we consider three major types of errors that may prevent a successful insertion: manufacturing error, sensing error, and manipulation error. We demonstrate how different constraints can be introduced and formatted into the automated design framework presented in this work. The overall design problem is modeled as a multi-stage constrained optimization problem. 

We improve and evaluate the design from two sub-processes. First, given the possible errors, will the insertion be successful? Second, given the errors, after a successful insertion, what is the maximum flexibility for the {\em coupled} joints? The insertion process is a pass-or-fail type of test. Under a linear assumption, some design changes may lead to failed insertion attempts, especially when the current {\em feasible} design is near the {\em boundary} of a feasible region. Such design changes need to be avoided if possible. We select the opposite changes to increase the chance of insertion success or to reduce the possible failure under disturbance. The second process evaluating the maximum flexibility after insertion, which we refer as {\em stability}. This is a quantitative measure. A better design should reduce such possible rotations. For example, Figure~\ref{fig:stuck} shows how our optimized design can improve the chance for insertion success and withheld external forces after insertion. 

In this work, we assume a small change of design affects the quantitative measures of insertion and stability linearly. For stability analysis, it has been shown that a linear estimation of the error between assembled blocks is reasonably accurate~\cite{Lensgraf20}. For insertion analysis, all the edges on the socket as well as the force being applied on the inserting joint are linear. Following this linear assumption, our derived designs show advantages compared to non-optimized designs. For example, in Figure~\ref{fig:stuck}, we show that in the simulation, the optimized design (right) does not get stuck, but other non-optimized designs will (left and middle). 

We also assumed a point-edge contact between the insertion joint and the socket. This assumption is introduced to overcome the uncertainties in the manufacturing process. By assuming point-edge contact, the location of contact can still be known despite some manufacturing error, as long as the point-contact cannot be {\em overshadowed} by manufacturing error. In this work, the point-contact is introduced as a circular bump on the insertion joint. 

This work has some obvious shortcomings. First, the analysis is currently in 2D. Though the 2D analysis is shown to be effective, the actual joints need to be in 3D. How does the 2D designs project to 3D is not yet fully analyzed. We only present a few simple 3D projections in this work to show the initial results. Second, we did not consider the grasping errors that may exist between gripper and block. Though this error can be removed by designing special grippers and blocks to allow automated alignment. The linear assumption also lacks the theoretical foundations. Last but not least, the design process is built upon the improvement on a initial design. We did not show that the feasible design space is fully connected, and the proposed procedure also need to loop over different number edges and point-contact pairs. 

\subsection{Our contributions}

Our main contributions in this work are as follows: 
\begin{enumerate}
    \item We introduce a procedure that automates the design for given constraints; 
    \item We model the errors from different sources into the design process and generate practical designs; 
    \item We separate the design into different processes to balance different {\em objectives} in the design; 
    \item We show the optimized design is unlikely to get stuck even with non-adaptive strategies and with reasonable errors; 
\end{enumerate}

\subsection{Related work}

Assembly with joinery has a long history in construction, such as the dovetail and mortise and tenon are used in carpentry around the world. Most such approaches were engineered and carried out entirely by humans. The major difference between the joinery based assembly and the brick-cement based assembly is the lack of {\em glues} and bonding materials~\cite{zwerger2012wood}, which make the assembly process reversible. Recently, it has been hypothesized that the backbones of Theropod dinosaurs interlocked to provide support for the extremely large body mass~\cite{woodruff2016}.

Reconfigurable assembly relying on geometric constraints has been well studied~\cite{song2012recursive, song2017reconfigurable, songcofifab, fu2015computational, Wang-2018-DESIA}. Different interlocking designs has been proposed in the past~\cite{Yao:2017:IDS:3068851.3054740}. 

Autonomous assembly with robots has also been explored before. Andres {\em et al.} created ROCCO to grasp and lay-down blocks for assembly~\cite{andres1994first}, and the similar system was later adapted by Balaguer {\em et al.}~\cite{balaguer1996site}. More systems have been developed recently, such as the system developed by Helm {\em et al.}~\cite{helm2012mobile} and by Giftthaler {\em et al.}~\cite{giftthaler2017mobile}. More novel construction approaches with drones and 3D printing technologies were also explored recently~\cite{willmann2012aerial, augugliaro2014flight, lindsey2011construction, augugliaro2013building, keating2017toward, winsun3dprint}. 

To allow the automation of the joinery-based assembly, there are some compromises one need to make. The joinery design needs to be less complicated for the ease of manufacturing and mass production. Also, the mechanism proposed by the joineries should be simple enough so that even with little adaptability, the robot devices can successfully supply the motions needed for the connection of joineries. Naturally, the joinery-based assembly extends from the idea of modular robots and assembly~\cite{rus2001crystalline, white2005three, romanishin2013m, daudelin2017integrated}. Recently, Schweikardt {\em et al.}~\cite{schweikardt2006roblocks} proposed an educational kit for robotic construction, inspired by LEGO. 

The simplest joint design is insertion-based, where the assembly process is just repeatedly apply translations to blocks while relying on the geometries of the blocks to secure the constructed structure. The classic peg-in-hole problem models this insertion strategy, which was studied by Lozano-Perez {\em et al.}~\cite{Lozano-Perez1984}. Building on the back-chaining approach, many similar systems were developed to study the insertion-based assembly approach~\cite{Mason86, Bruyninckx95, ZhangZOH04}. However, most study focus on the strategy for the insertion rather than the design of the joints. 

The automated design process presented in this work relies on numerical optimization, and a discretization of change in contacts during motion. In 2002, Balkcom {\em et al.} analyzed the possible motions of rigid objects under multiple contacts and given forces~\cite{Balkcom2002c}. In Moll's thesis~\cite{Moll2002}, a similar discretization approach was adapted to study the contacts between hands and objects. In this work, we also assume the linear edges in the socket design, thus using an linear approximation of more complex shapes. Linear approximation can greatly simplify the optimization process and find good results locally. For example, Berenson used linear approximation to study the effect of force applied to flexible objects~\cite{Berenson2013-deformable}. 

In this work, we adapted the assumption of point-surface contact, making the problem an extension of the caging problem~\cite{RimonBlack96, Rodriguez2010}. However, as the caging studies instances of contacts between fingers and objects, the design process introduced in this work also considers the changes and consequences of different contacts and forces. 



\section{Design process: from insertion to stability}

We show that the complete design procedure alternates between the optimizations for insertion and stability until the design cannot be improved. In our analysis, we assume there can be errors in the sensing and manipulation, resulting in wrong initial insertion configuration. Such insertion errors, however, will be assumed to be upper bounded, and the insertion process should be successful for any configuration within the error range for a good design. The manufacturing error is introduced as a uniform scale of the socket, even though the fabrication can appear both on socket and on joint. 

There are two aspects of the joint design we can change. First, the socket edges, and second the contact-points on the peg. During the insertion process, we only consider the design changes {\em initiated} by rotating socket edges. The contact point changes are being analyzed in the stability process.

\subsection{Insertion}

We use a point-edge contact model for our analysis, which leads to a relatively tolerant definition of the success criteria for insertion: contacts-points on the peg is contacting the predefined edges on the socket, and no undefined contacts are made. Due to manufacturing errors, there may exist contact-points on the peg not contacting the predefined edge on the socket even when the insertion is successful. So, the {\em goal state} is the maximum subset of the matching contact-points and socket edges. 

In order for the insertion to be successful, the joint must be able to reach the goal state from any initial configuration within the bounded errors. The design should admit a {\em path connected} space containing all possible initial configurations and goal state. This idea is very similar to the back-chaining used in the peg-in-hole problem, except we are confirming whether a design can make all back-chaining routes connected. 

We discretize the insertion process into a set of {\em contact modes} and the transitions among them. Given that the contact points on the joint is defined as $c_i, i=1\ldots, n$, and edges on the socket as $e_j, j=1,\ldots, m$, define the contact pair as a tuple $p_{i, j} = (c_i, e_j)$, and define a contact mode as a unique collection of contact pairs $M(i) = \{p_{a, b}, p_{c, d}, \ldots\}$. The goal state then can be described by a contact mode $M(g)$. Note that more than one contact mode can be in the goal state.


Though there can be different configurations correspond to the same contact mode, these configuration are {\em connected}. As long as the transition among different contact modes are connected, the overall transition process from initial configurations to the goal state is connected. However, there are contact modes where only a finitely many configurations can correspond to the given contact mode. The planar analysis is directly related to the study of immobilization and caging.

\begin{lemma}
Given a socket with linear edges, a set of points (peg), and a contact mode $M(k)$, if $|M(k)| > 2$, $|\cup e_j| < 3$ for all $e_j\in p_{i, j}\in M(k)$, and the edges are not parallel, then there are only finite configurations corresponding to the contact mode. 
\end{lemma}

We omit the proof of this and the following lemmas due to space constraint. The details of the proof can be found in Appendix. 




\begin{lemma}
Given a socket with non-parallel linear edges and a set of points (peg) that are not co-linear, if two valid contact modes $M(i)$ and $M(j)$ both have size at least $n+1$, then for any trajectory connecting $M(i)$ to $M(j)$ must go through a different contact mode $M(k)$ where $|M(k)| \leq n$.
\end{lemma}


\begin{lemma}
Given a socket of non-parallel linear edges and a set of not col-linear points (peg), for any valid contact mode $M(i) = \{p_{a,b}, p_{c,d}, p_{e,f}\}$, any combination of $p_{a,b}$, $p_{c,d}$ and $p_{e,f}$ is a valid contact mode.
\end{lemma}

\begin{figure}
\begin{center}
\includegraphics[width=3in]{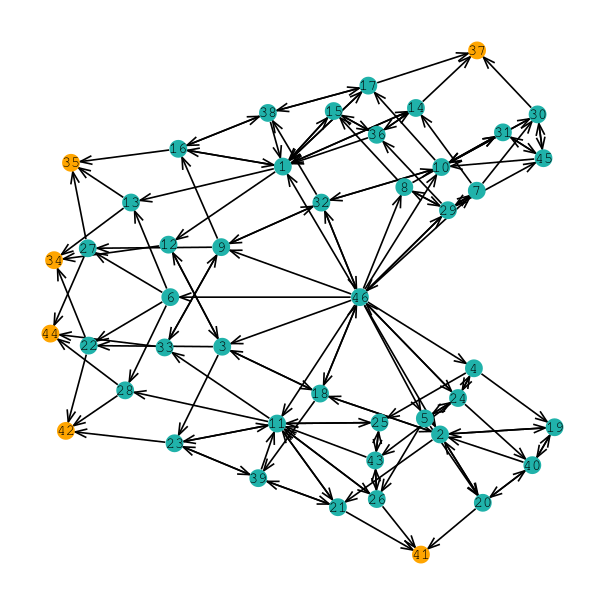}
\end{center}
\caption{A insertion graph of a five-point and five-edge peg-socket joint. }
\label{fig:insertion_graph}
\end{figure}

The insertion process analysis can be briefly described as follows. Given an initial design input, we first identify all valid contact modes using methods inspired by back-chaining. We then construct a directed graph $G_I = (V, E)$ where each vertex is a contact mode, as shown in Figure~\ref{fig:insertion_graph}. An edge $e(u\rightarrow v)\in E, u, v\in V$ is created if contact mode $u$ can transfer to $v$ following the insertion direction. Once the graph is constructed, we can identify {\em sinks} as the contact mode that have only incoming edges. A sink is not a desired sink if it is not in {\em goal state}. There can be multiple sinks if they are all in the goal state, and can transit among each other. 

When an insertion {\em fails}, i.e. the $G_I$ has undesired sinks, we need to remove these undesired sinks. We find the socket edge rotation directions that can remove the undesired sinks through a local trail and error process. We denote such rotation directions as the {\em gradient} direction that will increase the chance of success of insertion. If no changes of design can be introduced to remove the undesired sink, we declare the design is a failure. 

Below are some important details related to insertion analysis. First, we used back-chaining to detect all the valid contact modes, starting from the goal-states. Second, we adapted similar force analysis used in~\cite{Balkcom2002b} to detect the possible motion directions and connections to adjacent contact modes. We considered the pushing force is always along the tip of the block in this work, but the analysis and procedure hold if the force is along other directions. We also only considered the contact modes that has at most one contact pair difference. The contact mode transitions that creates or removes more than one contact pairs can either always be decomposed as a sequence of adjacent contact mode transitions, or be associated with special geometry or special force directions that are not along the insertion directions.

\begin{lemma}
Given a socket of non-parallel linear edges and a set of points (peg) that are not co-linear and not degenerate, and two valid contact modes $M(i)$ and $M(j)$. If $|M(i)| - |M(j)| > 1$ or $|M(i)| > n$ and $|M(j)| > n$, any trajectory connecting $M(i)$ to $M(j)$ can be written as a sequence of contact modes $M_1$, $M_2$, $\ldots$, $M_l$, where $M_1 \subseteq M(i)$ and $M_l \subseteq M(j)$, so that $|M_k| \leq n$ and $||M_k| - |M_{k+1}|| \leq 1$, $k= \{1, 2, \ldots, l-1\}$. 
\end{lemma}

Here, by degenerate, we mean that some edge lengths on the socket is the same as some distances between adjacent contact points on the joint. We omit the proof of the lemma, the detail of which can be found in our technical report~\cite{iros2021}. One of the most important steps in the analysis is to include the possible contact modes that are created due to the initial errors. Any identified contact mode will be part of the graph $G_I$.


\subsection{Stability}


The stability analysis is only initiated when $G_I$ has no undesired sink, i.e. can insert successfully. There are two factors that affect the stability of the peg after insertion. First, how much the peg can rotate before leaving the goal state, and second, how large the external forces need to be to push the peg away from the goal state. 

In the goal state, we want to minimize the maximum possible rotation angle to increase the stability, and we do so by sliding the contact points along the socket edges toward the direction that can reduce this angle. The correct sliding direction can be obtained through exploration. 

When the possible rotation angle is reduced for the goal state, we also need to consider how much force is needed to move the peg out of the goal states. If the force cone is very small, a small misaligned force can move the peg out of the goal state, making the rotation angle bound irrelevant. If multiple goal states are connected to each other, we will not analyze the forces that move the peg among these goal states. For example, for $G_I$ shown in Figure~\ref{fig:insertion_graph}, once the two undesired sinks ($37$ and $41$) are removed, and all remaining sinks are inter-connected, the force cone we want to maximize is bounded by the minimum force that can move the peg away from any of these sinks.

\subsection{Complete procedure}

The complete procedure that combines the above insertion analysis and stability analysis can then be described in Procedure~\ref{procedure:optdesign}. 

\begin{myprocedure}
\caption{Optimize design}
\label{procedure:optdesign}
\textbf{Input: Initial positions for all $p_i$ and $e_j$}; $\epsilon > 0$; $\Delta x, \Delta\theta$\\
Compute all possible initial contact modes for the given $\Delta x$ and $\Delta\theta$; add to $G_I$\\
\While{Improvements can be made} {
	Construct $G_I$ for the current design;\\
	Derive directions of edge rotation that may break insertion;\\
	\If {$G_I$ has undesired sinks or disconnected} {
		Rotate socket edges to remove the undesired sinks or connect $G_I$; \\
		\If {cannot remove undesired sink or connect $G_I$} {
			\textbf{break};
		} 
	}
	Find point moving directions that can reduce maximum rotation and increase force-cone; \\
	\While{The move of $p$ can improve stability not $\epsilon$ close to end-points of corresponding edge $e$} {
		Update best design;
	}
}
return the best design; 
\end{myprocedure}



The procedure iterates between the validation of insertion and improvements of stability, and stops when no progress can be made. There is one major flaw to this procedure: it needs an initial input of design. This means, the $n$ and $m$ is fixed for the entire run of this procedure. What is more, the corresponding edges and points are also fixed during the procedure. So, to find the actual best design overall, we need to loop over different combinations of $m$ and $n$, as well as different corresponding relations when $m \neq n$. Luckily, we do not need to consider the case where $m > n+1$ or $n > m+1$, as such case would result in redundancy. We also do not need to consider the case where $n > 6$, as that is also redundant base-on the analysis of planar caging and immobilization. 

\section{Validation, simulation, and experiments}

\begin{figure}[t]
\begin{center}
\includegraphics[width=2.3in]{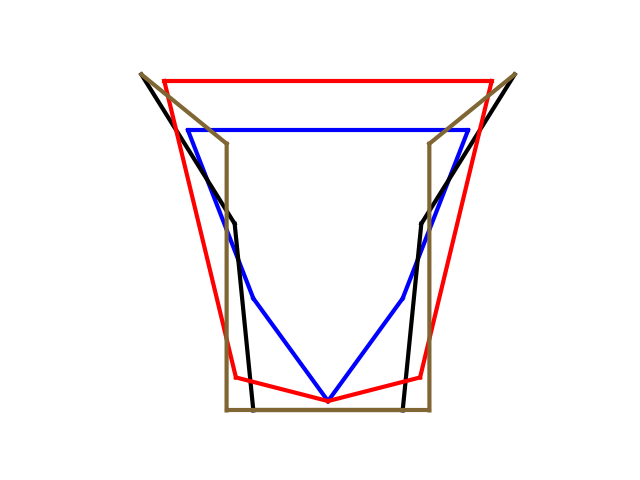}
\end{center}
\caption{Optimized peg-socket joint design (red and brown) and given peg-socket joint (blue and black) with five peg points and five socket edge. }
\label{fig:best_5_5_joint}
\end{figure}

\begin{figure}
\begin{center}
\begin{subfigure}[t]{0.23\textwidth}
\begin{center}
\includegraphics[height=1.1in]{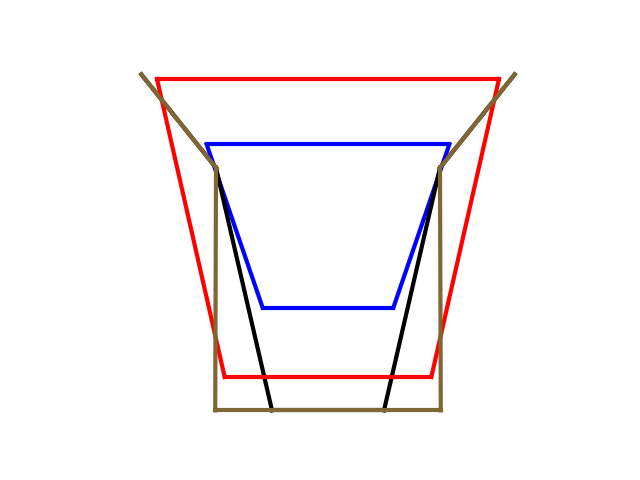}
\end{center}
\caption{Case for $m = 5, n = 4$. }
\label{fig:best_4-5_joint}
\end{subfigure}
\begin{subfigure}[t]{0.23\textwidth}
\begin{center}
\includegraphics[height=1.1in]{figures/5_5_final.png}
\end{center}
\caption{Case for $m = 5, n = 5$. }
\label{fig:best_5-5_joint}
\end{subfigure}
\begin{subfigure}[t]{0.23\textwidth}
\begin{center}
\includegraphics[height=1.1in]{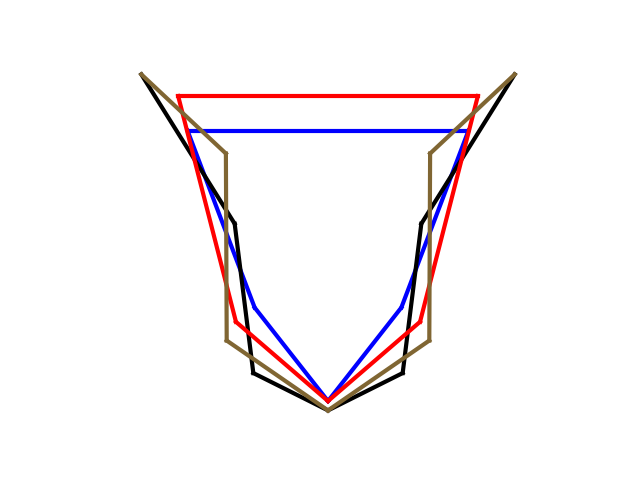}
\end{center}
\caption{Case for $m = 6, n = 5$. }
\label{fig:best_5-6_joint}
\end{subfigure}
\begin{subfigure}[t]{0.23\textwidth}
\begin{center}
\includegraphics[height=1.1in]{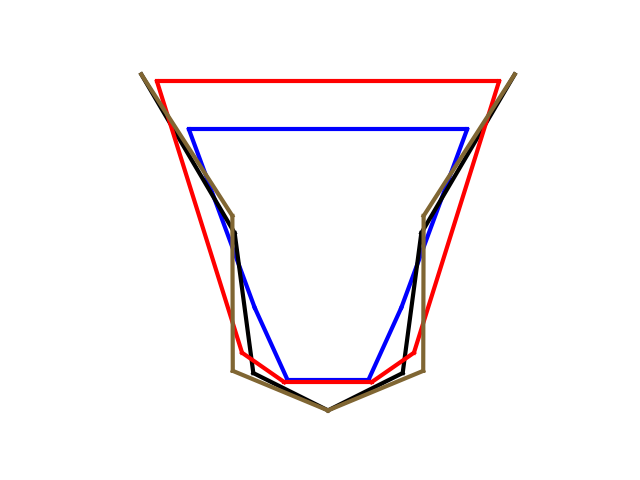}
\end{center}
\caption{Case for $m = 6, n = 6$.  }
\label{fig:best_6-6_joint}
\end{subfigure}
\caption{The initial (blue and black) and optimized (red and brown) designs  for joints with different $m$ and $n$. }
\label{fig:best_joint}
\end{center}
\end{figure}

We first tested the procedure in 2D, and compared different numbers of $m$ and $n$. The best designs are shown in Figure~\ref{fig:best_5_5_joint} and~\ref{fig:best_joint}. According to data from Table~\ref{table:m-n-opt}, the $4$ points $5$ edges joint design produces the least possible rotation in the goal states, followed closely by the design with $5$ points and $5$ edges. However, the force cone that maintains peg within the socket for $5-5$ design is considerably larger than $4-5$ design. 
 
\begin{table*}[h!]
  \begin{center}
    \caption{Stability after insertion measured by the max possible rotation angle, before and after the optimization for design.}
    \label{table:m-n-opt}
    \begin{tabular}{|c |c | c | c| c|}
    \hline
      \textbf{$n-m$} & \textbf{ Joint design } & \textbf{Successful insertion} & \textbf{Max-rotations for goal states}& \textbf{Stable force cone}\\
      \hline
       \multirow{3}{*}{4-5} & Initial input design & True & 0.025383698073 & [-1.331592653, 1.331592653]\\
      \cline{2-5}
     & Insertion optimized design & True & 0.025383698073 & [-1.331592653, 1.331592653]\\
      \cline{2-5}
     & Stability optimized design& True & 0.021741391097 & [-1.501592653, 1.501592653] \\
      \hline
     \multirow{3}{*}{5-5} & Initial input design & False & null & null\\
      \cline{2-5}
     & Insertion optimized design & True & 0.035501817173 & [-1.6615926535, 1.6615926535]\\
      \cline{2-5}
     & Stability optimized design& True & 0.024392557685 & [-1.7615926535, 1.7615926535] \\
      \hline
    \multirow{3}{*}{5-6} & Initial input design & False & null & null\\
      \cline{2-5}
     & Insertion optimized design & True & 0.07316540118 & [-1.3815926535, 1.3815926535]\\
      \cline{2-5}
     & Stability optimized design& True & 0.061805692421 & [-1.4915926535, 1.4915926535]\\
      \hline
     \multirow{3}{*}{6-6} & Initial input design & False & null & null\\
      \cline{2-5}
     & Insertion optimized design & True & 0.06201851152 & [-1.3715926535, 1.3715926535]\\
      \cline{2-5}
     & Stability optimized design& True & 0.041606704720 & [-1.591592653, 1.591592653] \\
      \hline
    \end{tabular}
  \end{center}
\end{table*}


We also tested the design in bullet physics simulation on the chance of successful insertion as we as the stability after insertion. By applying a unidirectional insertion force, we show that our design does not get stuck regardless of the initial insertion condition, while small perturbations of designs may result in premature lock, i.e. jammed. The possible rotations are also being tested in simulation by applying forces near the boundary of the computed force cone for our optimized design. Our optimized design can remain in the socket while other non-optimized designs will be moved out of the socket. 

\begin{figure}
\begin{center}
\includegraphics[width=2.5in]{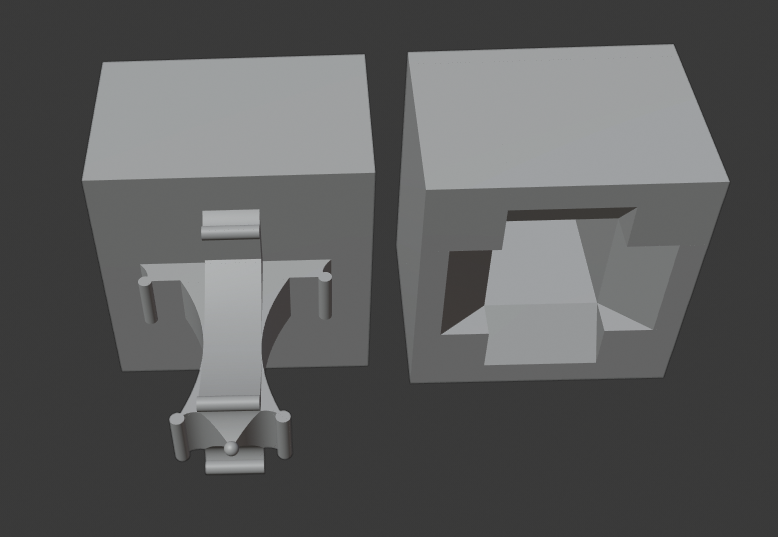}
\caption{A 3D joint model using our analysis projected to 3D at a $90$ degree separation.}
\label{fig:3d_model}
\end{center}
\end{figure}

\begin{figure*}[t]
\begin{center}
\begin{subfigure}[t]{0.15\textwidth}
\begin{center}
    \includegraphics[width=1.0in]{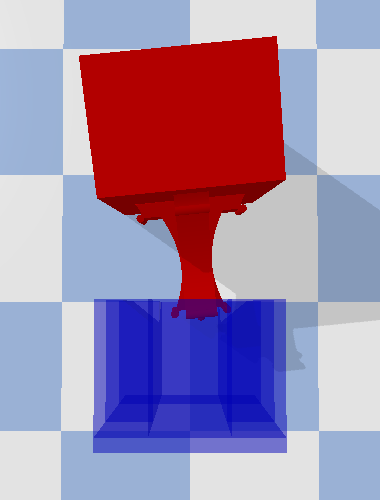}
\end{center}
\end{subfigure}
\begin{subfigure}[t]{0.15\textwidth}
\begin{center}
    \includegraphics[width=1.0in]{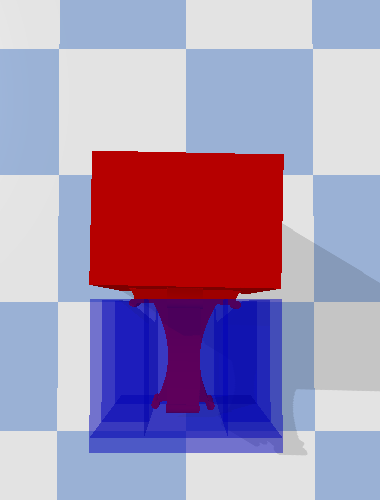}
\end{center}
\end{subfigure}
\begin{subfigure}[t]{0.15\textwidth}
\begin{center}
    \includegraphics[width=1.0in]{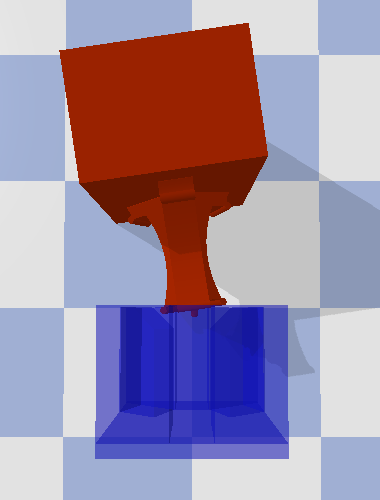}
\end{center}
\end{subfigure}
\begin{subfigure}[t]{0.15\textwidth}
\begin{center}
    \includegraphics[width=1.0in]{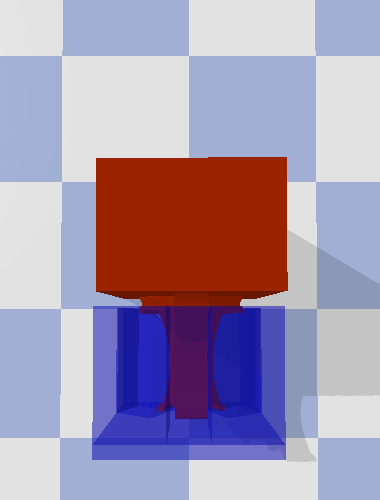}
\end{center}
\end{subfigure}
\begin{subfigure}[t]{0.15\textwidth}
\begin{center}
    \includegraphics[width=1.0in]{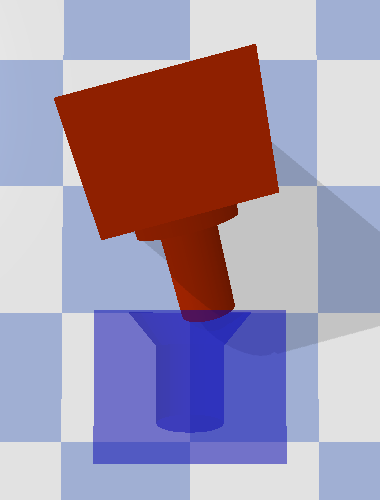}
\end{center}
\end{subfigure}
\begin{subfigure}[t]{0.15\textwidth}
\begin{center}
    \includegraphics[width=1.0in]{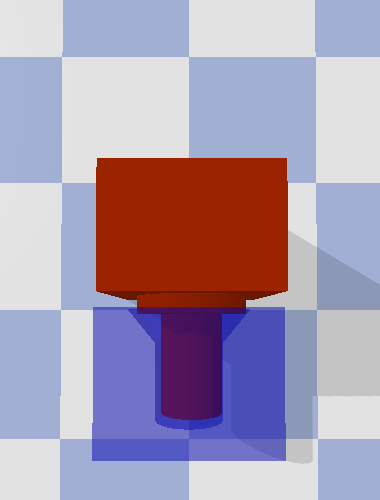}
\end{center}
\end{subfigure}
\caption{From left to right, each pair of images shows the initial insertion configuration of a 3D peg and the final insertion result in simulation. From left to right, the 3D peg is projected from 2D design to 3D with $90$, $120$, and $0$ degree separation. The $0$ degree separation projection results in a cylindrical peg. }
\label{fig:3d_simulation}
\end{center}
\end{figure*}

We further projects the design into 3D. However, in this initial work, we can only conduct a few naive projection schemes: a $90$, $120$ and $0$ degree separation projection. A $90$ degree separation projection is shown in Figure~\ref{fig:3d_model}. We tested the 3D design, and found that the design is easy to insert and stable after insertion when not too much yaw rotation is introduced, as shown in Figure~\ref{fig:3d_simulation}. When large yaw error exists at initial configuration, the $90$ and $120$ degree projection can be less successful during insertion. A cylindrical projection may be more effective for insertion, as shown on the right of Figure~\ref{fig:3d_simulation}, but can be less stable afterwards with respect to yaw rotations. So, we will further analyze the 3D projection schemes for better error tolerance analysis, and balance between the insertion and stability error tolerances. 

\section{Conclusions and future work}

In this work, we introduce an automated procedure to design a peg-in-hole type of joint. The procedure finds the {\em best} design in the existence of sensing, manipulation, and manufacturing errors. We assumed point-edge contact in 2D, and separated the analysis for insertion and after insertion stability. The procedure alternate the optimization between insertion and after insertion stability, until no improvements can be found. 

The design is validated in simulation. The results show that our design do have benefits, and increases the chance of success under uncertainties. We also started to look into the extension to 3D, and showed preliminary results. One of the major future work directions is the exploration and analysis for 3D projects. We will also use the derived designs to enable the stable assembly of larger structures.
\section{Appendix}

\subsection{Proof of Lemma 1}
Given a socket with linear edges, a set of points (peg), and a contact mode $M(k)$, if $|M(k)| > 2$, $|\cup e_j| < 3$ for all $e_j\in p_{i, j}\in M(k)$, and the edges are not parallel, then there are only finite configurations corresponding to the contact mode. 

\noindent
\begin{proof}
Given any valid configuration with $n$ points $p_1,..,p_n$ contacting on $m$ edges $e_1,..,e_m$, let’s move any two contact points $p_i$ and $p_j$ on their contact edges $e_i$ and $e_j$. 

If $e_i$ = $e_j$ or $e_j$ is parallel to $e_i$, the peg is sliding without rotating during $p_i$ and $p_j$ moving, that is, all the trajectories of other contact points would be parallel to $e_i$. Notice there exists at least one point $p_k$ contacting edge $e_k$ non-parallel to $e_i$, the trajectory of $p_k$ has only one intersection with its contact edge $e_k$, which gives only one valid configuration to maintain its point-edge contact relations.

If $e_i$ is non-parallel to $e_j$, the peg is continuously rotating during $p_i$ and $p_j$ moving. The trajectories of other contact points would be curves, which have only finite intersections with their linear contact edges. 
\end{proof}

\subsection{Proof of Lemma 2}
Given a socket with non-parallel linear edges and a set of points (peg) that are not co-linear, if two valid contact modes $M(i)$ and $M(j)$ both have size at least $n+1$, then for any trajectory connecting $M(i)$ to $M(j)$ must go through a different contact mode $M(k)$ where $|M(k)| \leq n$.

\noindent
\begin{proof}
Choose any configuration $q_i$ of $M(i)$ and $q_j$ of $M(j)$, the configuration set of any trajectory from $q_i$ to $q_j$ is a compact set $[q_i,  q_j]$ which includes infinite configurations. Notice $n \geq 2$, $n+1$ contact pairs give three or more points contacting on at least two non-parallel edges. From \em{Lemma1}, we have only finite configurations of $M(i)$ and $M(j)$. Thus from $q_i$ to $q_j$, the peg must go through infinite configurations of $M(k)$ with $n$ or less contact pairs.
\end{proof}

\subsection{Proof of Lemma 2}
Given a socket of non-parallel linear edges and a set of not col-linear points (peg), for any valid contact mode $M(i) = \{p_{a,b}, p_{c,d}, p_{e,f}\}$, any combination of $p_{a,b}$, $p_{c,d}$ and $p_{e,f}$ is a valid contact mode.

\noindent
\begin{proof}
Given a valid configuration of $M(i)$, for any two contact pairs $p_{a,b}$ and $p_{c,d}$, move $c_a$ and $c_c$ on their contact edges $e_b$ and $e_d$. From {\em{Lemma1}}, we have only finite valid configurations of $M(i)$, so the third contact pair $p_{e,f}$ must break its contact relation, i.e., $c_e$ must leave or penetrate $e_f$ during the two points moving. 
As the trajectory of $c_e$ would be a line non-parallel to $c_f$ or a curve open towards rotation point, which is inside the socket, a part of the trajectory must be inside the socket, which means there always exists a moving direction $d_k$ for the $c_a$ and $c_c$ that makes $c_e$ leave $e_f$.  Moving $c_a$ and $c_c$ along $d_k$, from {\em{Lemma2}}, after $c_e$ leaving $e_f$, there exist infinite valid configurations of $M(k) = \{p_{a,b}, p_{c,d}\}$ before other  points contacting. 
\end{proof}

\subsection{Proof of Lemma 3}
Given a socket of non-parallel linear edges and a set of points (peg) that are not co-linear, and two valid contact modes $M(i)$ and $M(j)$. If $|M(i)| - |M(j)| > 1$ or $|M(i)| > n$ and $|M(j)| > n$, any trajectory connecting $M(i)$ to $M(j)$ can be written as a sequence of contact modes $M_1$, $M_2$, $\ldots$, $M_l$, where $M_1 \subseteq M(i)$ and $M_l \subseteq M(j)$, so that $|M_k| \leq n$ and $||M_k| - |M_{k+1}|| \leq 1$, $k= \{1, 2, \ldots, l-1\}$. 


\begin{proof}
Let there be two Mode of Contacts $MoC(a)$ and $MoC(b)$ where a transition from $MoC(a)$ to $MoC(b)$ is possible, and the two modes differ by more than a single contact pair. Let there do not exist $MoC(c)$ where both $MoC(a)$ to $MoC(c)$ and $MoC(c)$ to $MoC(b)$ are possible. 

Without loss of generality, let $|MoC(b)| > |MoC(a)|$. and let $CP(i, s)$ and $CP(j, t)$ be the two contact pairs that cannot be introduced independently. There are two cases, 1) the peg can rotation even for the same contact location along the edge; 2), the orientation of the peg in $MoC(a)$ depends only on the contact location for the contact pairs, i.e. for the same contact location along the edges, no rotation is permitted. 

In the case 1, the only scenario is that there is only one contact pair in $MoC(a)$. In this case, as the orientation of the peg can change even when the contact pair does not change relative location, there does not exist design where $CP(i, s)$ and $CP(j, t)$ must make contact together unless some $d(p_i, p_j) = d(p^e_s, p^e_t)$, as any change of the orientation of the peg will change the relative distance between the pending contact pairs, preventing the simultaneous contact. 

In the case 2, as the socket has error, if the insertion angle is perfect aligned with the depth of the socket, no such $CP(i, s)$ and $CP(j, t)$ exist, as no two points are designed to contact the same edge, thus no simultaneous contacts. If the insertion angle is off by $\delta$, then unless the socket has non-uniform error, the rotation of the peg cannot permit multiple contact pairs to happen at the same time while they can also make contact when fully inserted into the no-error socket. 
\end{proof}

\subsection{Proof of Lemma 4}
Given a joint with sequence of point contacts $p_i, i=\{1, 2, \ldots, n\}$ with socket edges $e_j, j = \{1, 2, \ldots, m\}$, we can find all possible contact modes $M(\cdot)$, and construct partial order $\mathcal{S}$ on $M(\cdot)$ based on the maximum rotation for each contact mode. In the fully inserted configuration, if a single contact point $p_i$ slides along the contacting edge $e_j$ without creating a new contact mode, the partial order set $\mathcal{S}$ will remain the same. 

\begin{proof}
The partial order $\mathcal{S}$ is created based on the different possible rotations of different contact modes. If no new contact mode is created while the contact point is sliding along the corresponding contact edge, the partial order should remain the same as long as the sliding of contact points affect the rotation angle on all contact modes in the same gradient direction. 

Consider a single contact point, let us assume it slides towards one end of the corresponding edge and increase the distance with respect to the center of the joint. Then, all the rotations toward the sliding direction will meet an edge on the socket earlier if all the other contact points remain unchanged, i.e. rotation centers remain the same. A conflict will only be created when the rotation centers also slide along the edges and reduce the distance with respect to other contact points. But this conflicts the condition of only one contact point is moved. 
\end{proof}

\renewcommand*{\bibfont}{\small}
\printbibliography

\end{document}